\documentclass[letterpaper, 10 pt, conference]{ieeeconf}  

\IEEEoverridecommandlockouts                              

\usepackage{cite}
\usepackage{amsmath,amssymb,amsfonts}

\usepackage{algorithm}
\usepackage{algorithmicx}
\usepackage{algpseudocode}
\usepackage{mathtools}
\usepackage{amsmath}
\usepackage{mathrsfs}
\usepackage[dvipsnames]{xcolor}

\usepackage{hyperref}

\hypersetup{
    colorlinks=true,
    linkcolor=blue,
    filecolor=magenta,      
    urlcolor=cyan,
    pdftitle={Overleaf Example},
    pdfpagemode=FullScreen,
}

\title{\LARGE \bf
P2O-Calib: Camera-LiDAR Calibration Using \\ Point-Pair  Spatial Occlusion Relationship}

\author{Su Wang$^{1}$, Shini Zhang$^{1,2}$, Xuchong Qiu$^{1*}$
\thanks{$^{1}$authors with Bosch Corporate Research, Shanghai, China}%
\thanks{$^{2}$author with the Nanyang Technological University}%
\thanks{* the corresponding author 
{\tt\small xuchong.qiu@cn.bosch.com}}%
}

\newcommand{\etal}{\textit{et al.}}
\newcommand{\eg}{\emph{e.g.}}

\usepackage{graphicx}
\usepackage{textcomp}
\usepackage{xcolor}
\def\BibTeX{{\rm B\kern-.05em{\sc i\kern-.025em b}\kern-.08em
    T\kern-.1667em\lower.7ex\hbox{E}\kern-.125emX}}
    
\definecolor{DarkBlue}{rgb}{0.0, 0.5, 0.8}
\definecolor{Green}{rgb}{0.5, 1.0, 0}
\definecolor{yellow}{rgb}{1.0, 0.9, 0}

\usepackage{multirow}

\begin{document}

\maketitle
\thispagestyle{empty}
\pagestyle{empty}


\begin{abstract}
The accurate and robust calibration result of sensors is considered as an important building block to the follow-up research in the autonomous driving and robotics domain. The current works involving extrinsic calibration between 3D LiDARs and monocular cameras mainly focus on target-based and target-less methods. The target-based methods are often utilized offline because of restrictions, such as additional target design and target placement limits. The current target-less methods suffer from feature indeterminacy and feature mismatching in various environments. To alleviate these limitations, we propose a novel target-less calibration approach which is based on the 2D-3D edge point extraction using the occlusion relationship in 3D space. Based on the extracted 2D-3D point pairs, we further propose an occlusion-guided point-matching method that improves the calibration accuracy and reduces computation costs. To validate the effectiveness of our approach, we evaluate the method performance qualitatively and quantitatively on real images from the KITTI dataset. The results demonstrate that our method outperforms the existing target-less methods and achieves low error and high robustness that can contribute to the practical applications relying on high-quality Camera-LiDAR calibration.

\end{abstract}






\section{Introduction}
\label{intro}

In practical applications of autonomous driving and mobile robotics, such as detection and navigation,  multiple or even redundant sensors are often introduced into the system to ensure the reliability of environment understanding. Generally, cameras are considered as the first need, as they are low-cost and consistent with the perception of human eyes. Accordingly, the LiDARs can provide reliable 3D information about the environment, which the cameras cannot offer. To integrate these two heterogeneous sensors for better perception of environment, accurate extrinsic calibration is always of crucial importance.

\begin{figure}[t!]\label{fig_1}
    \centerline{\includegraphics[width=8.5cm]{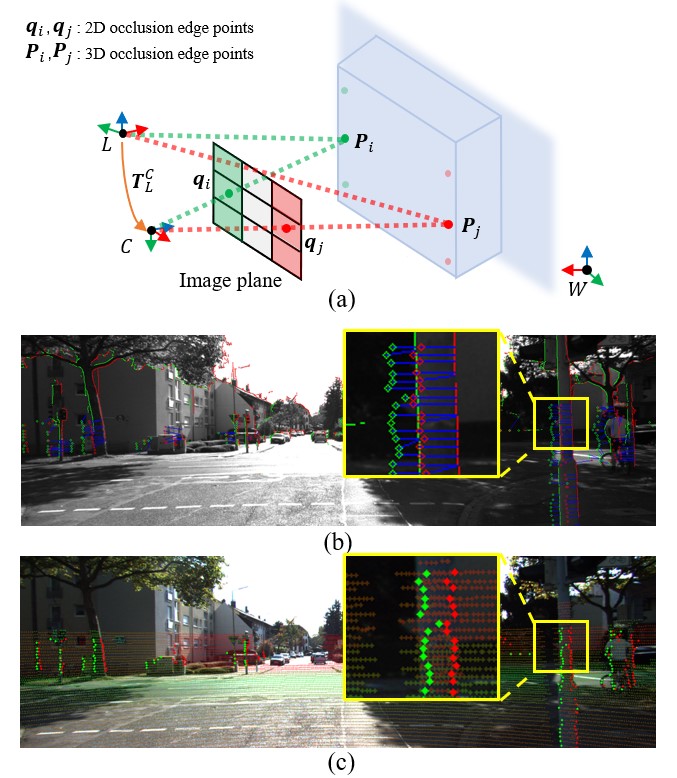}}
    \caption{Illustration of the proposed methods: (a) we propose to use the aligned 2D-3D occlusion edge point pairs (\eg, $(q_i, P_i)$, $(q_j, P_j)$) to estimate the extrinsic $T_{L}^{C}$ between camera $C$ and LiDAR $L$; (b) 2D-3D feature matching guided by occlusion directions. \textbf{Colored pixels} are occlusion edge feature points extracted from image, and \textbf{colored circles} are point cloud occlusion edge feature points from decalibrated LiDAR while \textcolor{green}{green} and \textcolor{red}{red} stand for the \textcolor{green}{left} and \textcolor{red}{right} occlusion directions respectively. \textcolor{blue}{Blue} lines stand for the data associations between features with the same occlusion direction (\eg, left occlusion edge); (c) projecting LiDAR point cloud on image with the estimated extrinsic.}
    \label{fig}
\end{figure}

Conventionally, prior work on Camera-LiDAR extrinsic calibration heavily relies on artificial targets, like checkerboard \cite{ICRA2012_single_shot} \cite{IROS2018_plane} \cite{ICRA2020_Planarity} \cite{arXiv2020_ACSC}, sphere \cite{ICRA2020_unified}, trihedron \cite{Sensors2013_trihedron} and etc. Such methods are usually carried out offline and in isolation from further processing, and they cannot rectify drift or vibration-induced calibration problems when the sensors are in use. Therefore, although these target-based methods can provide offline extrinsic calibration, online in-field calibration methods are more suitable for long-term applications in realistic environments.

Correspondingly, target-free methods eliminate the need for artificial targets, thereby simplifying online calibration. The features leveraged in target-free methods can be natural features~\cite{ICRA2013_LineBased}~\cite{ICRA2021_LineBased}~\cite{ICASSP2016_edge_align}~\cite{shi2020calibrcnn}~\cite{2021_pixel}~\cite{tafrishi2021line} or mutual information (MI)~\cite{AAAI2012_MI}~\cite{kang2020automatic}. However, the performance of target-free methods is usually restricted by the different nature between 2D-3D features~\cite{ICRA2013_LineBased}~\cite{ICRA2021_LineBased} or the limited generalization ability across different scenarios~\cite{shi2020calibrcnn}.       

To alleviate the aforementioned limitations, we propose a novel target-free extrinsic calibration method which uses the unified 2D-3D occlusion feature to achieve robust estimation in various scenarios. For feature extraction on camera images, our previous work~\cite{ECCV2020_P2ORM} introduces the concept of pixel-pair occlusion relationship and allows us to extract the image edges points with spatial occlusion information. In this work, we exploit the 3D occlusion edge extraction on LiDAR point clouds following the occlusion relationship definition, and develop a novel calibration approach using the extracted 2D-3D point pairs. Furthermore, the occlusion relationship offers helpful guidance to a more robust cross-modal feature association that can tolerate greater errors from the initial guess of Camera-LiDAR extrinsic parameters. Compared with prior work evaluated on the real-world dataset KITTI, the proposed method boosts both the accuracy and the generalization ability for Camera-LiDAR extrinsic calibration.



The main contributions of this work are:

\begin{itemize}
    \item A systematic Camera-LiDAR extrinsic calibration pipeline is designed on the basis of the occlusion relationship. It is accessible for online in-field calibration and free of external calibration targets.

    \item A novel point cloud feature extraction method and the corresponding 2D-3D edge features matching approach are proposed. The proposed method with directed occlusion information reduces feature registration mismatching and is robust to imperfect feature extractions.
    
    \item Experiments on both the synthetic and practical datasets validate the effectiveness of the proposed method, including ablation experiments and the examination for accuracy, robustness, and generalization ability. We will release our code and datasets.
    
\end{itemize}

The rest of this paper is organized as follows: related works towards extrinsic calibration between camera and LiDAR are reviewed in Sec.~\ref{sec:related_work}; Sec.~\ref{sec:method} elaborates on the method implemented in this work; the effectiveness of this work is qualitatively and quantitatively demonstrated in Sec.~\ref{sec:experiment}; the conclusion is drawn in Sec.~\ref{sec:conclusion} at last.



\begin{figure*}[ht]
    \centering
    \includegraphics[width=17.5cm]{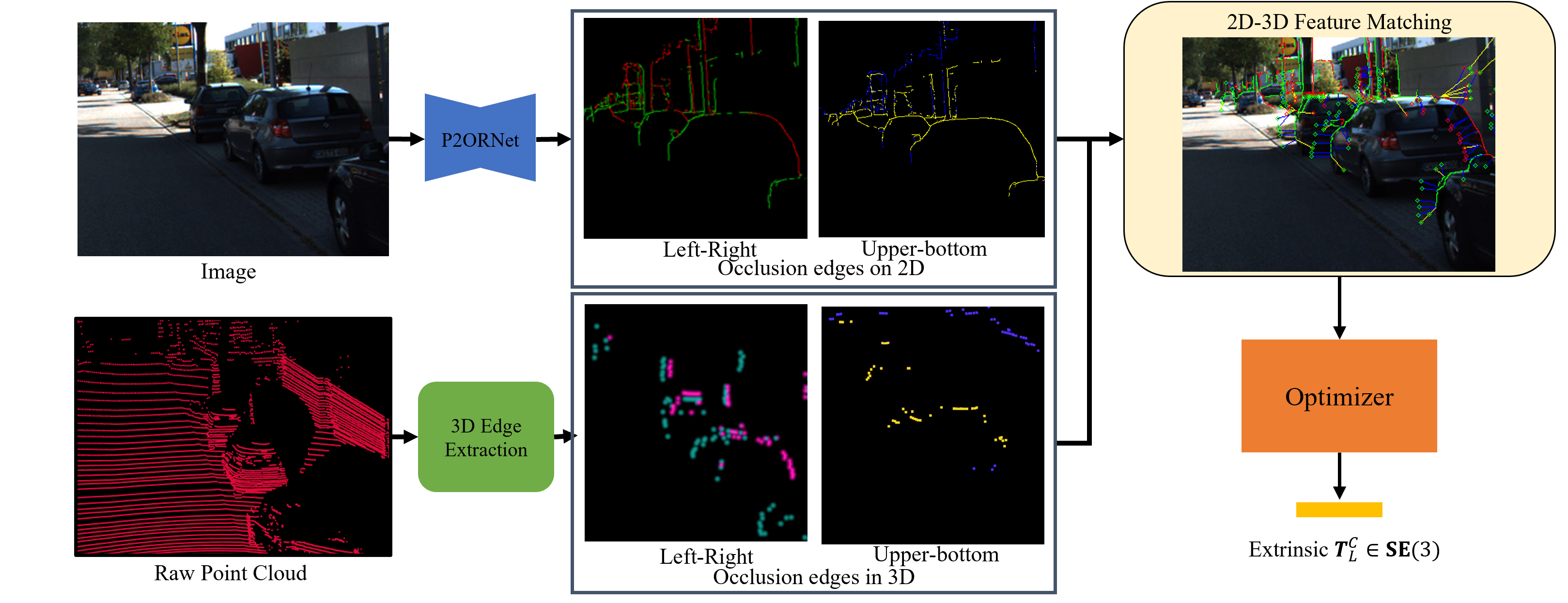}
    \caption{The architecture of the proposed method. Our method firstly extracts 2D-3D occlusion edge feature points from the input 2D-3D data and then applies the occlusion-guided feature matching to align them. Afterwards, an optimization is applied on the matched 2D-3D point pairs to recover the extrinsic between the camera and the LiDAR. The left and right  occlusion edge points are colored green and red while the upper and bottom occlusion edge points are colored yellow and blue. For clearness, only left-right occlusion edges are visualized in the 2D-3D Feature Matching module. Best viewed in a digital version.}
    \label{fig:flowchart}
\end{figure*}

\section{Related Works}
\label{sec:related_work}

The prevailing researches on Camera-LiDAR calibration were done in the following works. For the sake of systematic discussion, they are categorized into two main groups, target-based methods, and target-less methods.

\subsection{Target-Based Calibration}

Conventionally, target-based methods stand out as a classic solution to the Camera-LiDAR extrinsic calibration. On the planar level, the checkerboard \cite{arXiv2020_ACSC} \cite{IROS2018_plane} \cite{ICRA2020_Planarity} \cite{ICRA2012_single_shot} \cite{mirzaei2012_least-square} is widely utilized. Similar to the checkerboard, a planar board with four circular holes \cite{velas2014_3Dmarker} \cite{ITSC2017_velo2cam} can be detected by its rectangular and circular edges, not only applicable for RGB cameras but also for thermal cameras. On the spatial level, the sphere target \cite{ICRA2020_unified} is also considered, taking advantage of the sphere's symmetry. While calibration by the trihedron with the concave surface oriented towards the sensors \cite{Sensors2013_trihedron}, the spatial geometric characteristic of the trihedron also provides conditions for mmWave RADAR calibration \cite{ICRA2019_TUD} \cite{IEEE2021_TUD}. \cite{ICCV2017_boxes} describes a method based on ordinary boxes, whose structure is also a trihedron with convexity facing the sensors.

The widespread usage of calibration targets has specific features for detection in the environment. Inevitably, target-based methods introduce extra manufacturing tolerance and lead to laborious operations such as the target placement. In practice, there are some scenarios which are not suitable for the placement of targets. Notably, the complicated procedure and fixed calibration environment conflict with the requirement of online maintenance of extrinsic parameters for self-driving systems.

\subsection{Target-Less Calibration}

The target-less calibration is capable of operating before or during data collection, as well as identifying misalignment at different stages of online autonomous driving and robotics activity. Unlike target-based methods that revolve around targets, there are a variety of target-less ways to achieve sensor data alignment. For example, \cite{IEEE2007_point_corresp} use point correspondences manually selected from sensor data as the input of the PnP (Perspective-n-Point) algorithm to conduct extrinsic calibration. \cite{AAAI2012_MI} and \cite{ICRA2013_MI} propose probabilistic calibration approaches based on mutual information. However, the extrinsic parameters are sensitive to the variation of the crucial intensity information which usually leads to poor performance. \cite{IROS2018_motion} introduces a motion-based method to find the data correspondence between LiDAR and visual odometry. To resolve the scale ambiguity of camera, the initial extrinsic parameters are required from the LiDAR motions and camera motions. Compared with other methods, motion-based methods request more strict time synchronization across sensors. \cite{ICRA2013_LineBased} obtains the transformation between the coordinate frames by line feature registration in natural scenes. One limitation is that it works well in indoor scenes populated with a large number of line landmarks which are not easy to access in outdoor environments. \cite{ICRA2021_LineBased}, and its subsequent extension work \cite{arXiv2022_temporal} based on line features are similar to the proposed method. In \cite{sun2022atop}, a cross-modal network is proposed for object-level corresponding, and then the extrinsic is estimated by a heuristic algorithm. \cite{koide2023general} simplifies the 2D-3D feature alignment to image matching by leveraging learning-based feature. And Castorena \etal \cite{ICASSP2016_edge_align} also reports an algorithm via edge alignment but with additional depth maps and intensity images. Besides, end-to-end Camera-LiDAR calibration methods~\cite{schneider2017regnet}\cite{iyer2018calibnet}\cite{shi2020calibrcnn} extract 2D-3D features simultaneously and regress the $SE(3)$ extrinsic with learnable neural networks. The parameters of these networks converge to the training dataset statistics, and they are evaluated on the same dataset.

Our calibration pipeline aims to overcome the chaotic and redundant 2D edge features in~\cite{ICASSP2016_edge_align}\cite{ICRA2021_LineBased} by using the unified 2D-3D occlusion edge feature, and achieve high-quality online extrinsic calibration during in-field operations. Due to the robustness of our method, larger initial extrinsic parameters are also permitted for practical online calibration for the first time.


\section{Methodology}
\label{sec:method}


This section elaborates on the proposed Camera-LiDAR extrinsic calibration method. The pipeline is composed of three steps: First, a \textit{P2ORNet} neural network is trained for pixel-wise occlusion edge recognition on RGB image, and we extract the 3D occlusion edge from the given point cloud. Then we propose an occlusion-guided matching strategy between 2D pixels and 3D points for formulating a perspective-n-point(PnP) problem. Afterwards, the extrinsic calibration matrix is finally obtained through optimizing the point-to-line re-projection error. Fig.~\ref{fig:flowchart} sketches the proposed framework, and the following subsections will go into details of the definition of occlusion relationship and methods in each component.





\subsection{2D-3D Point Pair Definition}
\label{sec:pt_pair_def}

In this section, we formalize the definitions of 2D-3D occlusion edge and define the point pair between image 2D points and LiDAR 3D points using the occlusion relationship. 

\begin{figure}[h]
    \centering
    \includegraphics[width=0.4\paperwidth]{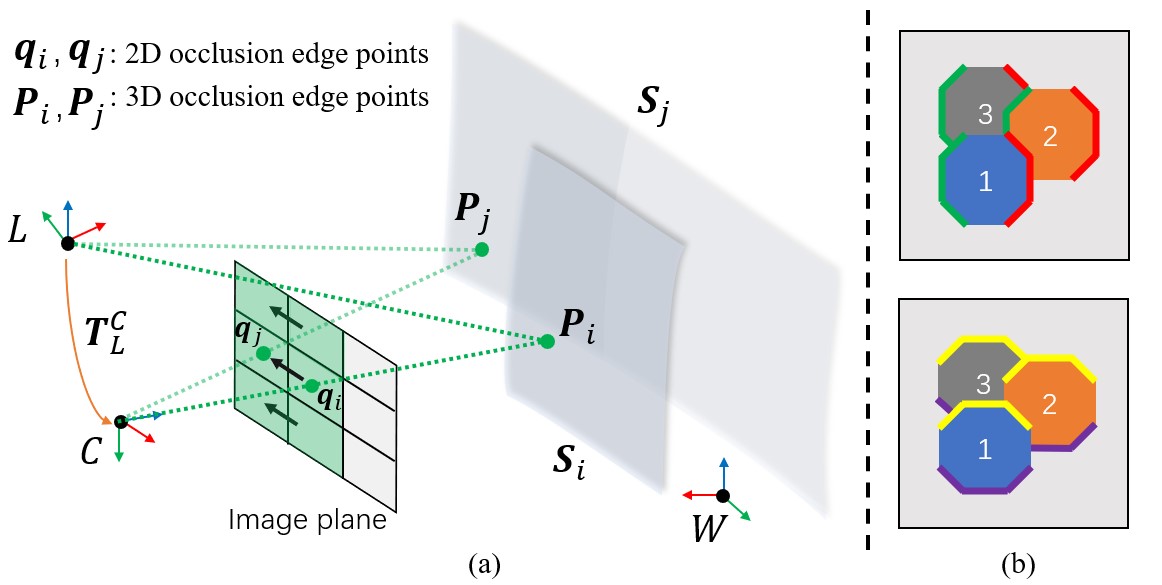}
    \caption{Illustration of 2D-3D occlusion edge point pair definitions: (a) image left occlusion edge region colored green and the corresponding 2D occlusion edge points (\eg, $q_i, q_j$) and 3D occlusion edge points (\eg, $P_i, P_j$), (b): image \textcolor{green}{left}-\textcolor{red}{right} occlusion edges and \textcolor{yellow}{upper}-\textcolor{blue}{bottom} occlusion edges in a scenario where there are three objects $1, 2, 3$ with occlusions.}
    \label{fig:2d_3d_def}
\end{figure}

As illustrated in Fig.~\ref{fig:2d_3d_def}(a), we consider a camera located at $C$ and a LiDAR located at $L$ who observe the spatial surfaces ($\textbf{S}_i, \textbf{S}_j$) of a 3D scene with reference to the world frame $W$. The $SE(3)$ transformation from LiDAR frame $L$ to camera frame $C$ is $\textbf{T}_L^C$, and the task of estimating $\textbf{T}_L^C$ is defined as extrinsic calibration in this work. We note $\textbf{q}_i, \textbf{q}_j$ are the camera image plane 2D projections (pixels) of 3D points $\textbf{P}_i, \textbf{P}_j$ which lie on the surfaces $\textbf{S}_i, \textbf{S}_j$ respectively. From the perspective of $C$ and $L$, $\textbf{S}_i$ comes in front of $\textbf{S}_j$, and the occlusion edges lies on the boundary region between $\textbf{S}_i$ and $\textbf{S}_j$ non-occluded part. On the left-side occlusion edge region, we note $\textbf{q}_i, \textbf{q}_j$ as 2D occlusion edge feature points and $\textbf{P}_i, \textbf{P}_j$ are 3D occlusion edge feature points. The black arrow pointing to $\textbf{q}_j$ indicates that $\textbf{S}_i$ occludes $\textbf{S}_j$ from right to left in 3D space on the edge region. By assigning such direction to each pixel pair, the occlusion edge feature points can be further divided into feature points with different directions (cf. Fig.~\ref{fig:2d_3d_def}(b)), for example, left occlusion edge feature points $\textbf{q}_i, \textbf{q}_j$ in this illustration. Starting from the occlusion relationship between surfaces, here we formally define that each 2D-3D point pair consists of a 3D feature point captured by LiDAR and a 2D feature point captured by the camera, \eg, $(\textbf{q}_i, \textbf{P}_i)$ and $(\textbf{q}_j, \textbf{P}_j)$, in this work. 

\subsection{Occlusion Edge Extraction in Images}

\subsubsection{Image connectivity-8 neighborhood}
For each image pixel $\textbf{q}_i$, it has 8 immediate neighbor pixels, which consist of 4 horizontal/vertical neighbors and 4 diagonal neighbors. One image pixel can therefore pair with its 8 neighbor pixels to form 8 pixel pairs.

\subsubsection{2D occlusion edge estimation} \label{sec:imageproc}
As introduced in Sec.~\ref{sec:pt_pair_def}, the surface-pair occlusion relationship in 3D can be represented as the pixel-pair occlusion relationship in 2D. For each valid image pixel pair $(\textbf{q}_i, \textbf{q}_j)$, neural network \textit{P2ORNet}~\cite{ECCV2020_P2ORM} classifies the occlusion relationship status among three possible three status: $\textbf{q}_i$ occludes $\textbf{q}_j$, $\textbf{q}_j$ occludes $\textbf{q}_i$ and no occlusion between $(\textbf{q}_i$ and $\textbf{q}_j)$. The estimated image occlusion edge lies on the image region where \textit{P2ORNet} predicts that occlusion exists between pixel pairs. Without loss of generality, we choose occlusion edge foreground pixels as the occlusion edge points in 2D (cf. Fig.~\ref{fig:flowchart}). In this work, we choose pixel pairs that are connected along the image's horizontal/vertical axis and generate left-right ($\mathcal{L}, \mathcal{R}$) and upper-bottom ($\mathcal{U}, \mathcal{B}$) occlusion edges from the prediction results. We note $\mathbb{Q}^{\mathcal{D}}$ image 2D occlusion edge feature point set with direction $\mathcal{D}$ while $\mathbf{q}^{\mathcal{D}}_{i}$ is the $i$th point:

\begin{align*}
    \mathbb{Q} &:= \{\mathbb{Q}^{\mathcal{D}} \mid \mathcal{D}=\mathcal{L}, \mathcal{R}, \mathcal{U}, \mathcal{B}\}, \\
{\rm where}\\
    \mathbb{Q}^{\mathcal{D}} &:= \{\mathbf{q}^{\mathcal{D}}_i \mid i = 1, 2, ..., N_Q \}.
\end{align*}

\subsubsection{Network Training}
The \textit{P2ORNet} is trained with a class-balanced cross-entropy loss~\cite{ECCV2020_P2ORM}, taking into account the low-class frequency of pixels on image occlusion edges (cf. Sec.~\ref{sec:p2ornet_lr_strategy} for more details). 

\subsection{LiDAR Feature Extraction} \label{sec_pcprocess}

LiDAR can naturally provides information for determining the differences in range among scans, making 3D occlusion edge feature extraction more intuitive than the 2D ones~\cite{ICRA2012_single_shot}\cite{shan2018lego}. The horizontal occlusion edge features are extracted by traversing each ring of the scan and marking the point as an occluded feature once they are discontinued to their neighboring beam on the same ring. The occlusion edge vertical features are extracted in the same way by traversing each column of the LiDAR and comparing to their vertical neighboring beams. To maintain one-to-one image-LiDAR feature correspondence, occlusion features from LiDAR are classified into four categories following the same logic in Sec.~\ref{sec:imageproc}.


Afterward, considering that points laying on the ground plane can hardly be occlusion points, RANSAC (Random Sample Consensus) \cite{RANSAC} is applied onto the raw point cloud to extract the coefficient of ground plane $\mathbf{S}_{pl}$ beforehand. After the feature extraction, the occlusion edge features close to $\mathbf{S}_{pl}$ are removed. Then a radius filter is applied to the feature clouds for reducing floating outliers.

Then, the 3D feature points set $\mathbb{P}$ from a single frame is defined as:

\begin{align*}
\label{equ_P_3d}
    \mathbb{P} &:= \{\mathbb{P}^{\mathcal{D}} \mid \mathcal{D}=\mathcal{L}, \mathcal{R}, \mathcal{U}, \mathcal{B}\} \\
{\rm where}\\
    \mathbb{P}^{\mathcal{D}} &:= \{ \mathbf{P}^{\mathcal{D}}_i \mid i = 1, 2, ..., N_P \}.
\end{align*}

An example of the extracted 3D feature point set is shown in the 3D edge extraction module in Fig. \ref{fig:flowchart}.

\subsection{Extrinsic Calibration}

\subsubsection{\textbf{Occlusion-guided feature matching (OGM)}} \label{sec_matching}

Our 2D-3D feature matching approach uses the occlusion direction to improve data association, i.e., the point cloud features only match the image features that have the same occlusion direction since the occlusion relationship of LiDAR point cloud features is unchanged after being projected. Thereby, four KD-trees are constructed separately from the four groups of image features $\mathbb{Q}^{\mathcal{L}}$, $\mathbb{Q}^{\mathcal{R}}$, $\mathbb{Q}^{\mathcal{U}}$, $\mathbb{Q}^{\mathcal{B}}$.

\begin{figure}[t]
    \centering
    \includegraphics[width=8.7cm]{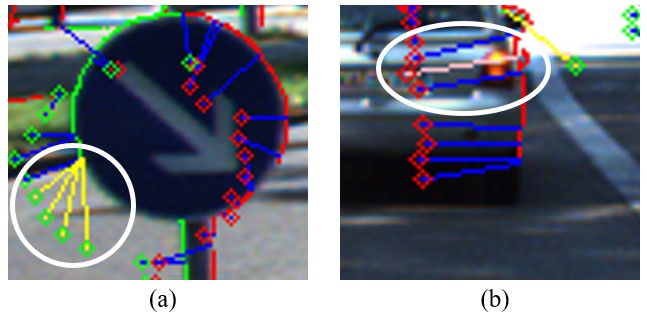}
    \caption{Examples of invalid feature pairs. Green/red rhombus: point cloud edge points, green/red lines: detected image occlusion edges, Blue lines: accepted matching, \textbf{yellow lines} in (a): matching with invalid angle, caused by missing image edge,  \textbf{pink line} in (b): the LiDAR feature is matched to a sharp image edge 
    curve and becomes invalid.} 
    \label{fig:mismatch}
\end{figure}

Then, we project each LiDAR feature $\mathbf{P}^\mathcal{D}_{i} \in \mathbb{P}^{\mathcal{D}}$ onto the camera frame via the un-calibrated initial extrinsic: $\mathbf{p}^{\mathcal{D}}_{i}=\pi (\widetilde{\mathbf{\xi}}^C_L, \mathbf{P}^\mathcal{D}_i,)$, where $\mathbf{\pi}$ is the projection function, and the initial extrinsic parameters $\widetilde{\xi}^C_L\in \mathfrak{se}(3)$ can be obtained from mechanical setup or the brutal force searching like \cite{2021_pixel}. For each projected point $\mathbf{p}^\mathcal{D}_i$, a reference pixel set with 8 nearest pixels is obtained from its corresponding KD-tree to establish a candidate 2D edge: 

\begin{align*}
\mathbb{L}_{i}^\mathcal{D}:=\{\mathbf{c}^{\mathcal{D}}_i, \mathbf{n}_i^\mathcal{D}\},
\end{align*}

\noindent in which $\mathbf{c}^\mathcal{D}_i$ is the geometric gravity center of the 8 points, $\mathbf{n}^\mathcal{D}_i$ is the eigenvector corresponding to the smallest eigenvalue of the 8-point set covariance matrix $\mathbf{M}_{cov,i}$.


In the next step, we filter out the outlier matching pairs using geometric information. First, the distances from any of the 8 nearest image features to $\mathbf{p}^\mathcal{D}_i$ should be smaller than a cutting distance $d_c$ in pixels, which is a changeable threshold during the optimization. The image feature may be partially absent owing to the limitation of the network as shown in Fig.~\ref{fig:mismatch}(a). Therefore, we threshold the angle between $\overrightarrow{\mathbf{c}^\mathcal{D}_i\mathbf{p}^\mathcal{D}_i}$ and $\mathbf{n}^\mathcal{D}_i$ so that $\arccos( \mathbf{n}^\mathcal{D}_i \cdot \overrightarrow{\mathbf{c}^\mathcal{D}_i\mathbf{p}^\mathcal{D}_i} / \| \overrightarrow{\mathbf{c}^\mathcal{D}_i\mathbf{p}^\mathcal{D}_i} \| )$ should smaller than a value $\lambda_{a}$. Besides, in case that matching $\mathbf{p}^\mathcal{D}_i$ to a sharp curve edge leads to inconsistency, the largest eigenvalue of $\mathbf{M}_{cov,i}$ should be $\lambda_{c}$ times larger than the second largest one (cf.~Fig.~\ref{fig:mismatch}(b)). 

\begin{figure}[t]
    \centering
    \includegraphics[width=8.7cm]{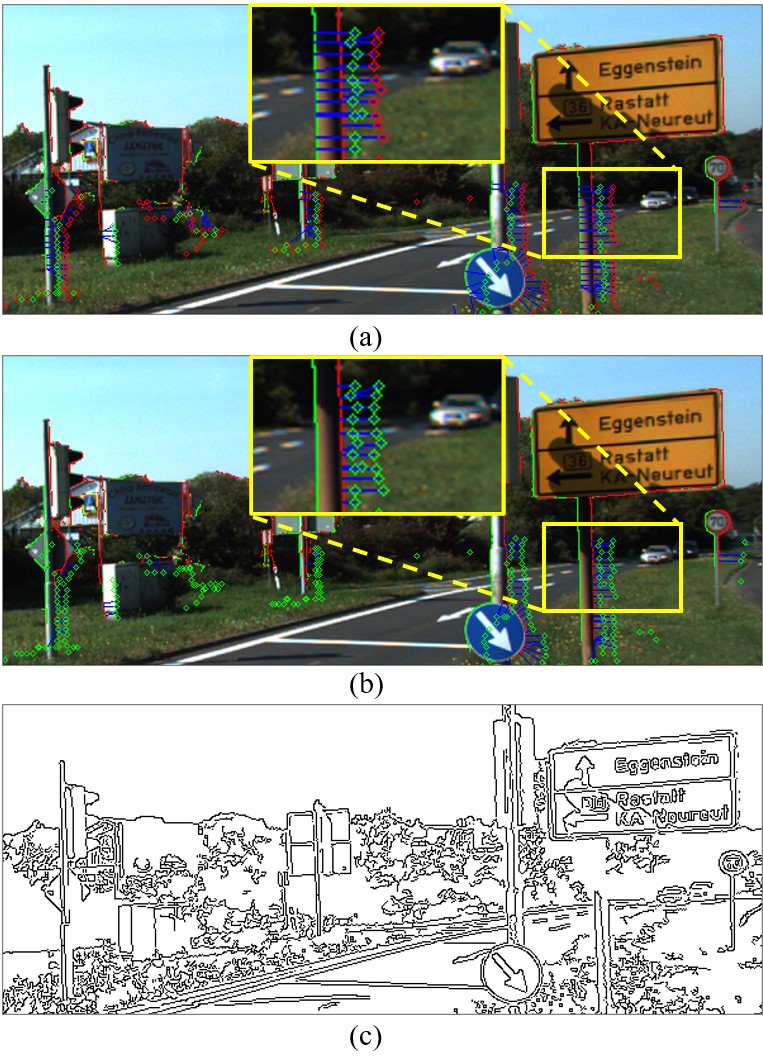}
    \caption{Comparison of feature extraction and matching: Figure (a) shows the proposed feature matching result. Image occlusion edges are associated with the point cloud occlusion edges with the same occlusion directions. In Figure (b), feature positions are the same as (a) but directly associated without occlusion guidance. In the zoomed window in (b), all the left and right point cloud features of the pole are incorrectly matched to the right image edges. Figure (c) shows image of noisy Canny edges which are not consistent with the nature of 3D edges on point clouds.}
    \label{fig:canny_comparison}
\end{figure}

An example of our matching result is shown in Fig. \ref{fig:canny_comparison}(a), while Fig. \ref{fig:canny_comparison}(b) is the matching between two identical features, but without the guidance of occlusion direction. It is obvious that the point features are prone to mismatching due to initial extrinsic error. Edges extracted solely using image gradient (\eg, Canny~\cite{canny1986computational}) are shown in Fig \ref{fig:canny_comparison}(c) as a comparison, with clearly denser and carry additional invalid information from 2D patterns.

\subsubsection{\textbf{Optimization}} \label{sec_optimization}

Given a number of associated 2D-3D features, the calibration problem can be solved as a PnP problem. Aiming to reduce the reprojection error from 3D occlusion edge point $\mathbf{P}^\mathcal{D}_i$ to 2D occlusion edge candidate $\mathbf{L}^\mathcal{D}_i$, the residual is defined as:

\begin{equation}
\label{equ_residual}
    \mathop{\arg\min}_{\xi} \sum^{\{ \mathcal{L}, \mathcal{R}, \mathcal{U}, \mathcal{B}\}}_{\mathcal{D}} \sum^{N_D}_{i}
\rho_i\left(\left\|f_r\left(\pi(\widetilde{\mathbf{\xi}}_\mathbf{L}^\mathbf{C},\mathbf{P}^\mathcal{D}_i), \mathbf{L}^\mathcal{D}_i   \right)\right\|^2\right),
\end{equation}

\noindent $f_r$ is the perpendicular distance from $p_i^d$ to $L$:

\begin{equation}
\label{equ_projection}
f_r\left( \mathbf{p}, \mathbf{L}\right) = \mathbf{n} \cdot \mathbf{\overrightarrow{\mathbf{c}\mathbf{p}}},
\end{equation}

\noindent where $N_D$ is the number of matched feature pairs in a specified operation iteration, $\rho$ refers to Huber kernel \cite{huber1992robust} that applied for balancing the weights of the outlier matches. And the Eqt. \ref{equ_residual} is optimized using the Levenberg-Marquardt method.

The optimization should be conducted for $\mathbf{n}_{opt}$ times with the re-associated features. For the first run of optimization, a large cutting distance $d_c$ is set for a more aggressive matching attempt and then generally reduced in the next iterations. The full optimization step is summarized in the Algorithm~\ref{alg_optimization}. The algorithm can also take accumulated feature pairs from multiple images as input, ensuring efficiency and adaptability.

\begin{algorithm}[t] \label{alg_optimization}
	\caption{Extrinsic optimization}     
	 \label{alg_optimization}      
	\begin{algorithmic}[1] 
	\Require Initial associated occlusion feature pairs with directions $\mathbb{M}=\{ \mathbf{L}^\mathcal{D}_i, \mathbf{P}^\mathcal{D}_i \mid i=1,2,...,N_m \}$
	\Require All features $\mathbb{Q}$, $\mathbb{P}$, \Comment{Sec. \ref{sec:imageproc}, Sec. \ref{sec_pcprocess}}
	\Require initial extrinsic parameters $\mathbf{\xi}^\mathcal{D}_i$, initial cutting distance       
    \Ensure extrinsic matrix $\mathbf{\widetilde{T}}_L^C \in SE(3)$  
    \Repeat
        \For {each $\{\mathbf{L}^\mathcal{D}, \mathbf{P}^\mathcal{D}\}$} in $\mathbb{M}$
            \State Extend residual block from $\{\mathbf{L}^\mathcal{D}, \mathbf{P}^\mathcal{D}\}$ \Comment{Eqt. \ref{equ_projection}}
        \EndFor
   \State Optimize $\mathbf{\widetilde{\xi}}_L^C$ \Comment{Eqt. \ref{equ_residual}}
   \State Update $d_c$ \Comment{Sec. \ref{sec_optimization}}
   \State Update $\mathbb{M}$ based on $\mathbb{P}, \mathbb{Q}, \mathbf{\widetilde{\xi}}_L^C$ \Comment{Sec. \ref{sec_matching}}
   \Until {reach maximum optimization times $n_{opt}$}
   \State\Return $\mathbf{\widetilde{T}}_L^C = {\rm \textbf{Exp}} \left( \mathbf{\widetilde{\xi}}_L^C \right)$ 
   \end{algorithmic}
\end{algorithm}

\section{Experiments}
\label{sec:experiment}

\subsection{General Experiment Setup}

\subsubsection{Ground Truth Preparation}

The model of \textit{P2ORNet} used for all the following tests is trained on the virtual KITTI2 (vKITTI2)~\cite{vkitti2} dataset. We use the depth images of vKITTI2 and the tools provided by~\cite{ECCV2020_P2ORM} to generate the occlusion relationship ground truths. In order to get rid of the trivial but noisy occlusion image features, e.g., features from leaves on the trees, occlusion labels belonging to these categories of objects are masked out in the training dataset.

The ground truth extrinsic $\mathbf{T}_{gt}$ is provided in both the synthetic and real-world datasets. To test on the decalibration cases, a disturbance transformation $\mathbf{T}_{e}$, which consists of the errors in both translation and rotation, is applied to the ground truth and, therefore, the initial guess is set to $\widetilde{\xi^\mathbf{C}_\mathbf{L}}=$\textbf{Log}$(\mathbf{T}_{gt}\mathbf{T}_{e})$. 

\subsubsection{Learning strategy of \textit{P2ORNet}}
\label{sec:p2ornet_lr_strategy}

We initialize the encoder model of the \textit{P2ORNet} with the weights of a ResNet-50 model~\cite{he2016deep} pre-trained on ImageNet~\cite{deng2009imagenet}, and the remaining layers with random values (as defined by the PyTorch default initialization). To train the network, we use the ADAM optimizer~\cite{kingma2014adam} with a learning rate $10^{-4}$ and divide it by 10 when half of the total training iterations (180,000 for the vKITTI2~\cite{vkitti2}) is reached. The input image size during training is $1242\times345$, and the mini-batch size is 4.

\subsubsection{Other implementation details} 
\label{sec:optim_details}

The least-square problem Eqt.~\ref{equ_residual} is solved using Ceres Solver \cite{Ceres2022}. Besides, some key variables are listed as followings: the initial cutting distance: $d_c=50$[pixel], the maximum number of optimizations: $n_{opt}=10$, angle threshold for feature matching: $\lambda_{a}=50$[deg], occlusion distance threshold: $\lambda_{occ}=40[cm]$.

\subsection{Experiment on the Synthetic Dataset}

\begin{figure}[ht]
    \centering
    \includegraphics[width=0.99\linewidth]{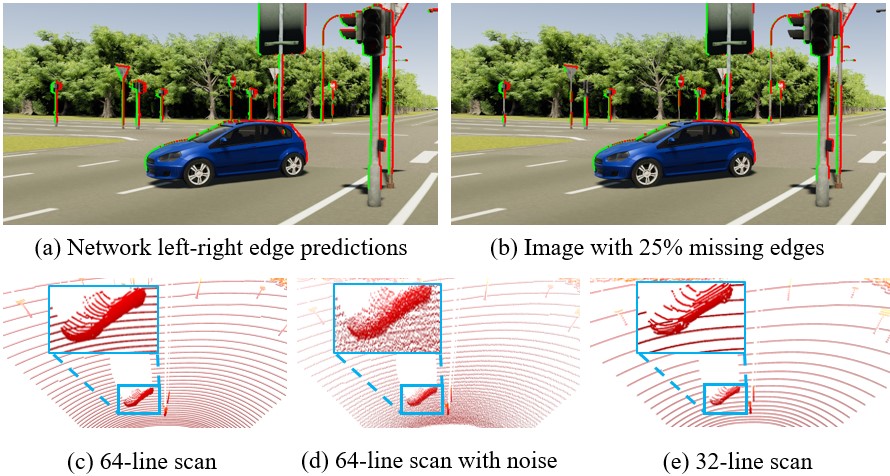}
    \caption{Examples of the synthetic data. }
    \label{fig_simu}
\end{figure}

\begin{table}[ht]
\caption{Result on the synthetic dataset with different configurations.}
\centering
  \begin{tabular}{ l||c c c|c c c}
    \hline
    \hline
    & \multicolumn{6}{c}{Calibration Result} \\
    \hline
     &
      \multicolumn{3}{c|}{Rotation Errors[deg]} &
      \multicolumn{3}{c}{Translation Errors[m]}
       \\
  \hline

    & roll & pitch & yaw & x & y & z \\
    
    \textit{test1} & 0.117 & 0.099 & 0.040 & 0.025 & 0.033 & 0.057 \\
    
    \textit{test2} & 0.170 & 0.134 & 0.062 & 0.067 & 0.048 & 0.081 \\
    
    \textit{test3} & 0.295 & 0.200 & 0.094 & 0.091 & 0.084 & 0.149 \\

    \textit{test4} & 0.108 & 0.130 & 0.056 & 0.034 & 0.076 & 0.104 \\

    \textit{test5} & 0.236 & 0.194 & 0.037 & 0.075 & 0.066 & 0.053\\

    \hline
  \end{tabular}
  \\
  \begin{tabular}{ l||c | c | c c}
    \hline
    \hline
     &\multicolumn{4}{c}{Experiment configuration}
    
       \\
  \hline
    &\multirow{2}{4em}{Lidar Type} & \multirow{2}{5em}{2D feature missing rate}
    & \multicolumn{2}{c}{Lidar noise} \\ & & & Range $\sigma_r$ & Angular $\sigma_{\alpha}$ \\
    \hline
    \textit{test1} & HDL64 & - & 0.02 & -  \\
    
    \textit{test2} & HDL64 & 25 & 0.02 & - \\
    
    \textit{test3} & HDL64 & 45 & 0.02 & -  \\

    \textit{test4} & HDL64 & - & 0.04 & 0.005  \\

    \textit{test5} & HDL32 & - & 0.02 & -  \\
    \hline
  \end{tabular}
  \label{table:1}
\end{table}

In the vKITTI2 dataset, LiDAR data is unavailable. To address this, we simulate partial LiDAR scans using depth images. Each LiDAR beam is parameterized by its direction angle expressed in polar coordinates and its range value is sampled from the patch of depth image that is traced along the direction angle. Realistic sensor noise is introduced with distance noises  $\mathbf{n}_r\sim\mathcal{N}(0, \sigma_r)$ and angular disturbances $\mathbf{n}_{\alpha}\sim\mathcal{N}(0, \sigma_{\alpha})$ to each beam. The pinhole camera model is used for both rendering and PnP solving. In this experiment, we rendered 64-line and 32-line LiDAR data using the manufacturing setup of Velodyne HDL64 and HDL32, respectively. The 2nd row of Fig. \ref{fig_simu} shows rendered scans with variant parameters.

To highlight the robustness of the proposed method, we focus on validating the calibration performance when the output of the network is not perfect. The raw edge detection result is utilized as the benchmark, as shown in Fig.~\ref{fig_simu}(a). To simulate the scenarios with imperfect network output, $x\%$ edge points are removed from it by masking the edges out with random circle patches, which is similar to the network performance in practice. Processed images with 25$\%$ edge missing rate are shown in Fig.~\ref{fig_simu}(b) as examples.

Tests were performed on sequences 1, 2, 6, 20 in vKITTI2, which contain driving scenes in urban area and on highways. Tab.~\ref{table:1} presents the mean absolute errors for each axis and the associated experiment parameters. The best result is achieved with $0\%$ missing edges, $\sqrt{\sigma_r}=0.02$m and $\sqrt{\sigma_{\alpha}}=0$ on the 64-line scans. It should be noted that our method can achieve relatively high accuracy when the missing edge percentage is close to $25\%$. Furthermore, our method is able to maintain a competitive performance even the noises are increased to $\sqrt{\sigma_r}=0.04$m and $\sqrt{\sigma_{\alpha}}=0.005$deg, which is almost double the Velodyne's manufacturing range measurement error. Compared to the results on 64-line scans, our method performance on 32-line scans slightly decreases.

\subsection{Experiments on the Real-world Dataset}

We then validate our method on the real-world KITTI dataset~\cite{kitti}, in which a Velodyne HDL-64 LiDAR and cameras are well calibrated, synchronized, and deskewed to satisfy the evaluation requirements of the proposed functions.

\begin{figure}[t]
    \centering
    \includegraphics[width=0.99\linewidth]{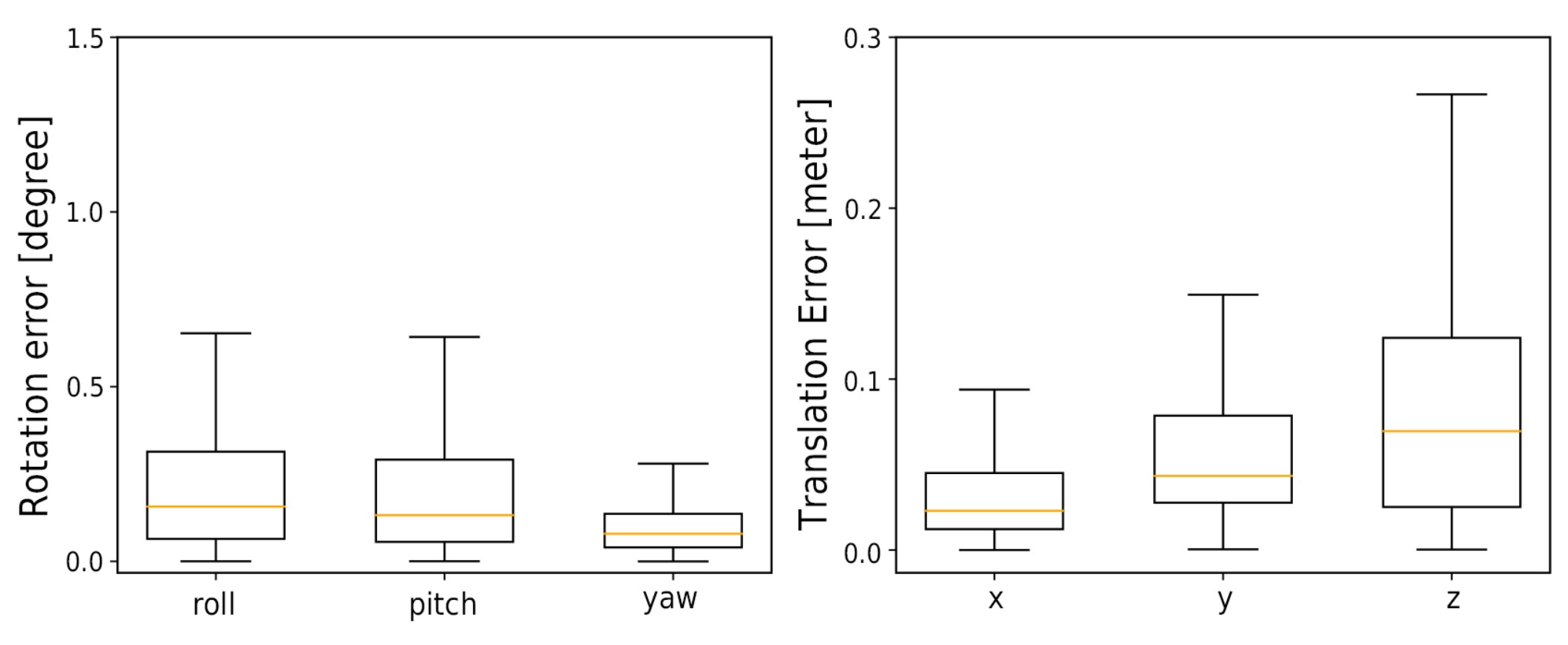}
    \caption{Performance of the method conducted on random rotation error ([-3, 3] degree), and translation error ([-0.15, 0.15] meter) in uniform distribution.}
    \label{fig_boxplot}
\end{figure}

\begin{figure*}[t]
    \centering
    \includegraphics[width=0.99\linewidth]{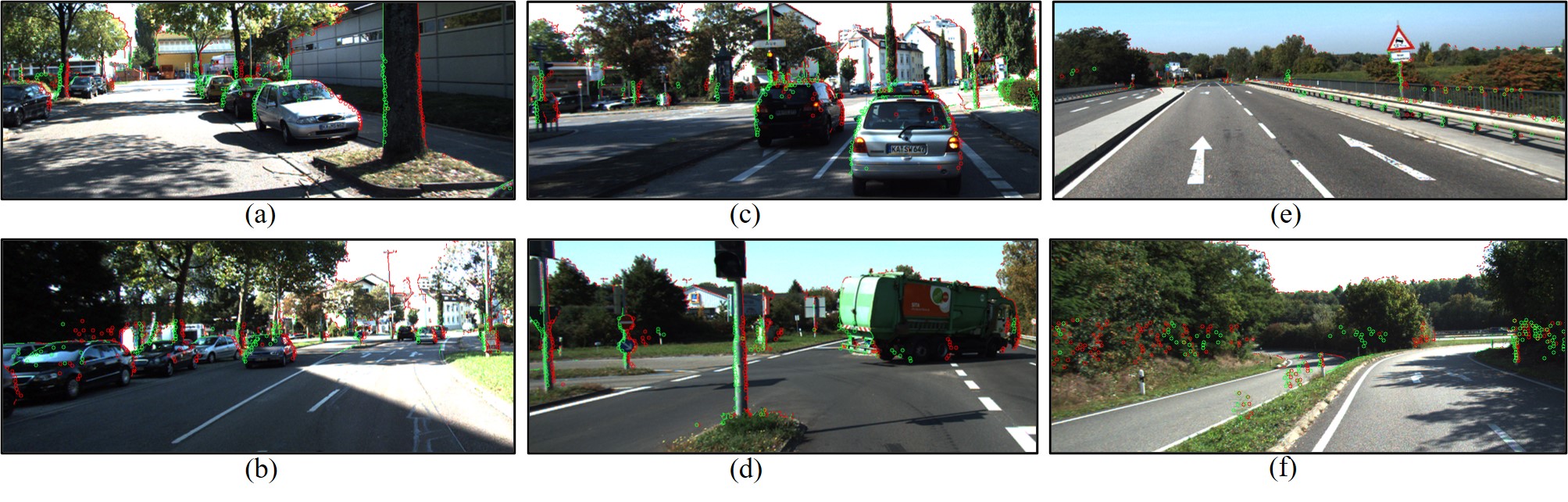}
    \caption{Experimental results on KITTI dataset. Thin edges represent 2D features while the circles stand for the 3D feature projected using estimated extrinsic transformation. All the figures are cropped for a clear view. The 1st and 2nd columns are examples that extrinsic is successfully calibrated. Note that the edge detection network suffers from overexposure and previously unknown vehicle in (c) and (d), respectively. And our method is able to handle the undesirable output. The 3rd column shows the failure cases since the edge detection result is not enough. For these two failure images, 3D point clouds are projected with ground truth only for visualization.}
    \label{fig_failure}
\end{figure*}

The experiments are run on the $09\_26$ sequences and the camera model is the rectified projection model of the KITTI raw dataset. The error $\mathbf{T}_{e}$ is set to a random transformation with a 2-degree rotation angle and 0.15-meter translation. To investigate the effectiveness of the proposed modules, we run experiments without each of them as the ablation study and offer the corresponding analysis. Table~\ref{table:2} shows the quantitative results. First, the missing of OGM (cf.~Sec.~\ref{sec_matching}) leads to failures in data association, and the results are therefore observably diverged. We then compare the results w/o upper-bottom occlusion edges (UB). The result indicates that the vertical edges positively affect the performance. However, it still can be seen from Fig. \ref{fig_boxplot} that the proposed method shows a greater uncertainty on the roll, pitch, and z-axis even with the vertical edges. We assume the reason is the lower vertical resolution of the LiDAR. Finally, to further verify the influence of the changeable feature distribution in different scenes, we supply another experiment on the sub-sequence $09\_26\_0096$ leveraging all the features collected from multiple frames in the sequence (MBS). Since abundant features from multiple frames provide a more stable confidence value, we conduct a grid search algorithm with 1 deg and 0.005m searching resolution in this test. Some of the calibration results are shown in Fig. \ref{fig_failure}, including successful results as well as failure cases in the dataset.



\begin{table}[t]
\centering
\caption{Mean Absolute Errors of Camera-LiDAR Calibration (Results from a single run with multiple frames indicated by *). The mean absolute errors (MAE) for rotations and translations are reported in degrees and meters, respectively. N/A is used when results are not provided by the authors. The symbol $\approx$ indicates that the value is approximated from the Roll-Pitch-Yaw (RPY) angles supplied by the authors..}
\begin{tabular}{l | c c | c c } 
 \hline
 Method  & $e_{\mathbf{R}}$ [deg] & $e_\mathbf{t}$ [m] & init \textbf{R} [deg] & init \textbf{t} [m] \\ [0.5ex] 
 \hline\hline
 RGGNet~\cite{yuan2020rggnet} &  1.180  & 0.088 & N/A  & N/A\\
 line-based~\cite{ICRA2021_LineBased} & $\approx$ 0.388 & N/A & 1.00 & 0.05\\
 \hline
 Ours w/o OGM & 2.799 & 0.32 & 2.00 & 0.15 \\
 Ours w/o UB & 0.32 & 0.142 & 2.00 & 0.15 \\
 Ours & 0.297 & 0.129 & 2.00 & 0.15 \\
 \hline
 Ours w/ MBS * & 0.284 & 0.102 & 10.00 & 0.15\\ [1ex]
 \hline
\end{tabular}
\label{table:2}
\end{table}

We also compare our algorithm with a number of existing calibration methods, which provide evaluations on the KITTI dataset, including a conventional \textit{line-based} method \cite{ICRA2013_LineBased} and learning-based methods \cite{yuan2020rggnet}. For a fair comparison, the learning-based methods are tested on different sequences other than the training sequence in KITTI. The results are shown in Table. \ref{table:2} as well. Our method with and without MBS outperforms the other methods in rotation. The proposed method with MBS also shows the second-best accuracy in translation among all the methods. Despite the best translation estimation performance, RGGNet shows significantly larger rotation errors than the others. Relying on traditional feature extraction and searching-based optimization, the \textit{line-based} method only provides results within a small range of initial error (1.0 deg rotation and 5 cm translation error). 

\begin{figure}[t]
    \centering
    \includegraphics[width=8cm]{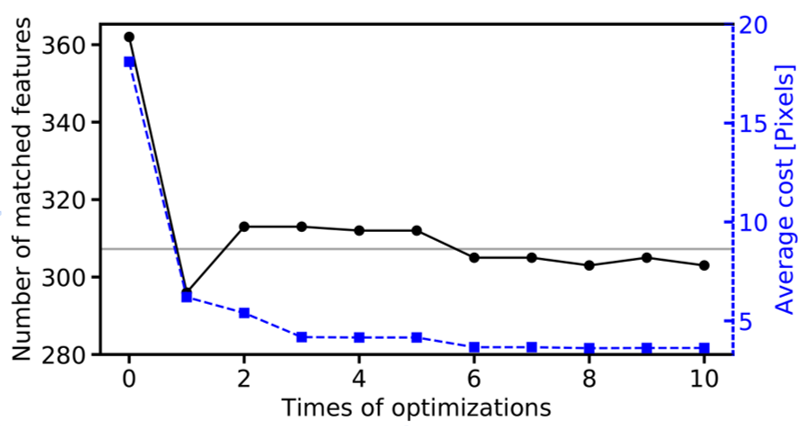}
    \caption{Changes of the feature number and average cost during convergence.}
    \label{fig:convergence}
\end{figure}

Finally, the scene depicted in Fig.~\ref{fig_1} is taken as an example for qualitatively validating the convergence performance. The changes in both the number of associated features and the means of residual over 10 iterations are shown in Fig~\ref{fig:convergence}. Because of the loose initial cutting distance threshold $d_c$, the solver starts with a high proportion of outliers and an average cost of over 18 pixels. After one cycle of optimization and feature re-association, both the cost and the number of feature pairs are reduced swiftly and then generally converge during several iterations.




\section{Conclusion}
\label{sec:conclusion}

In this paper, a targetless Camera-LiDAR calibration approach is proposed. Based on the occlusion relationship concept, we introduce a robust 2D-3D data association method that relies on natural 3D edge features in the scene. Extracting the 3D occlusion information in the 2D image offers us the chance to generate well-organized image edge features and align them with LiDAR point cloud features in an easy way. Moreover, the performance of our calibration method is robust to partial missing of 2D or 3D features. In contrast to the end-to-end calibration methods, the proposed method extracts image edges with the neural network but explicitly builds the feature pairs and calculates extrinsic via the nonlinear optimization. This hybrid architecture reduces the influence of statistical biases and cross-dataset domain gaps. Experiments on real-world data demonstrate that our method consistently outperforms the state of the art for target-less Camera-LiDAR calibration even if decalibrated initial values are given.   






\bibliographystyle{ieeetr}
\bibliography{ieeeconf}

\begin{thebibliography}{10}

\bibitem{ICRA2012_single_shot}
A.~Geiger, F.~Moosmann, O.~Car, and B.~Schuster, ``Automatic camera and range
  sensor calibration using a single shot,'' in {\em 2012 IEEE International
  Conference on Robotics and Automation}, pp.~3936--3943, 2012.

\bibitem{IROS2018_plane}
L.~Zhou, Z.~Li, and M.~Kaess, ``Automatic extrinsic calibration of a camera and
  a 3d lidar using line and plane correspondences,'' in {\em 2018 IEEE/RSJ
  International Conference on Intelligent Robots and Systems (IROS)},
  pp.~5562--5569, 2018.

\bibitem{ICRA2020_Planarity}
G.~Koo, J.~Kang, B.~Jang, and N.~Doh, ``Analytic plane covariances construction
  for precise planarity-based extrinsic calibration of camera and lidar,'' in
  {\em 2020 IEEE International Conference on Robotics and Automation (ICRA)},
  pp.~6042--6048, 2020.

\bibitem{arXiv2020_ACSC}
J.~Cui, J.~Niu, Z.~Ouyang, Y.~He, and D.~Liu, ``Acsc: Automatic calibration for
  non-repetitive scanning solid-state lidar and camera systems,'' {\em arXiv
  preprint arXiv:2011.08516}, 2020.

\bibitem{ICRA2020_unified}
J.~Kümmerle and T.~Kühner, ``Unified intrinsic and extrinsic camera and lidar
  calibration under uncertainties,'' in {\em 2020 IEEE International Conference
  on Robotics and Automation (ICRA)}, pp.~6028--6034, 2020.

\bibitem{Sensors2013_trihedron}
X.~Gong, Y.~Lin, and J.~Liu, ``3d lidar-camera extrinsic calibration using an
  arbitrary trihedron,'' {\em Sensors}, vol.~13, no.~2, pp.~1902--1918, 2013.

\bibitem{ICRA2013_LineBased}
P.~Moghadam, M.~Bosse, and R.~Zlot, ``Line-based extrinsic calibration of range
  and image sensors,'' in {\em 2013 IEEE International Conference on Robotics
  and Automation}, pp.~3685--3691, 2013.

\bibitem{ICRA2021_LineBased}
X.~Zhang, S.~Zhu, S.~Guo, J.~Li, and H.~Liu, ``Line-based automatic extrinsic
  calibration of lidar and camera,'' in {\em 2021 IEEE International Conference
  on Robotics and Automation (ICRA)}, pp.~9347--9353, 2021.

\bibitem{ICASSP2016_edge_align}
J.~Castorena, U.~S. Kamilov, and P.~T. Boufounos, ``Autocalibration of lidar
  and optical cameras via edge alignment,'' in {\em 2016 IEEE International
  Conference on Acoustics, Speech and Signal Processing (ICASSP)},
  pp.~2862--2866, 2016.

\bibitem{shi2020calibrcnn}
J.~Shi, Z.~Zhu, J.~Zhang, R.~Liu, Z.~Wang, S.~Chen, and H.~Liu, ``Calibrcnn:
  Calibrating camera and lidar by recurrent convolutional neural network and
  geometric constraints,'' in {\em 2020 IEEE/RSJ International Conference on
  Intelligent Robots and Systems (IROS)}, pp.~10197--10202, IEEE, 2020.

\bibitem{2021_pixel}
C.~Yuan, X.~Liu, X.~Hong, and F.~Zhang, ``Pixel-level extrinsic self
  calibration of high resolution lidar and camera in targetless environments,''
  {\em IEEE Robotics and Automation Letters}, vol.~6, no.~4, pp.~7517--7524,
  2021.

\bibitem{tafrishi2021line}
S.~A. Tafrishi, X.~Dai, and V.~E. Kandjani, ``Line--circle--square (lcs): A
  multilayered geometric filter for edge-based detection,'' {\em Robotics and
  Autonomous Systems}, vol.~137, p.~103732, 2021.

\bibitem{AAAI2012_MI}
G.~Pandey, J.~R. McBride, S.~Savarese, and R.~M. Eustice, ``Automatic
  targetless extrinsic calibration of a 3d lidar and camera by maximizing
  mutual information,'' in {\em Twenty-Sixth AAAI Conference on Artificial
  Intelligence}, 2012.

\bibitem{kang2020automatic}
J.~Kang and N.~L. Doh, ``Automatic targetless camera--lidar calibration by
  aligning edge with gaussian mixture model,'' {\em Journal of Field Robotics},
  vol.~37, no.~1, pp.~158--179, 2020.

\bibitem{ECCV2020_P2ORM}
X.~Qiu, Y.~Xiao, C.~Wang, and R.~Marlet, ``Pixel-pair occlusion relationship
  map (p2orm): formulation, inference and application,'' in {\em European
  Conference on Computer Vision}, pp.~690--708, Springer, 2020.

\bibitem{mirzaei2012_least-square}
F.~M. Mirzaei, D.~G. Kottas, and S.~I. Roumeliotis, ``3d lidar--camera
  intrinsic and extrinsic calibration: Identifiability and analytical
  least-squares-based initialization,'' {\em The International Journal of
  Robotics Research}, vol.~31, no.~4, pp.~452--467, 2012.

\bibitem{velas2014_3Dmarker}
M.~Vel'as, M.~{\v{S}}pan{\v{e}}l, Z.~Materna, and A.~Herout, ``Calibration of
  rgb camera with velodyne lidar,'' 2014.

\bibitem{ITSC2017_velo2cam}
C.~Guindel, J.~Beltrán, D.~Martín, and F.~García, ``Automatic extrinsic
  calibration for lidar-stereo vehicle sensor setups,'' in {\em 2017 IEEE 20th
  International Conference on Intelligent Transportation Systems (ITSC)},
  pp.~1--6, 2017.

\bibitem{ICRA2019_TUD}
J.~Domhof, J.~F. Kooij, and D.~M. Gavrila, ``An extrinsic calibration tool for
  radar, camera and lidar,'' in {\em 2019 International Conference on Robotics
  and Automation (ICRA)}, pp.~8107--8113, IEEE, 2019.

\bibitem{IEEE2021_TUD}
J.~Domhof, J.~F. Kooij, and D.~M. Gavrila, ``A joint extrinsic calibration tool
  for radar, camera and lidar,'' {\em IEEE Transactions on Intelligent
  Vehicles}, vol.~6, no.~3, pp.~571--582, 2021.

\bibitem{ICCV2017_boxes}
Z.~Pusztai and L.~Hajder, ``Accurate calibration of lidar-camera systems using
  ordinary boxes,'' in {\em Proceedings of the IEEE International Conference on
  Computer Vision (ICCV) Workshops}, Oct 2017.

\bibitem{IEEE2007_point_corresp}
D.~Scaramuzza, A.~Harati, and R.~Siegwart, ``Extrinsic self calibration of a
  camera and a 3d laser range finder from natural scenes,'' in {\em 2007
  IEEE/RSJ International Conference on Intelligent Robots and Systems},
  pp.~4164--4169, IEEE, 2007.

\bibitem{ICRA2013_MI}
Z.~Taylor and J.~Nieto, ``Automatic calibration of lidar and camera images
  using normalized mutual information,'' in {\em Robotics and Automation
  (ICRA), 2013 IEEE International Conference on}, Citeseer, 2013.

\bibitem{IROS2018_motion}
R.~Ishikawa, T.~Oishi, and K.~Ikeuchi, ``Lidar and camera calibration using
  motions estimated by sensor fusion odometry,'' in {\em 2018 IEEE/RSJ
  International Conference on Intelligent Robots and Systems (IROS)},
  pp.~7342--7349, 2018.

\bibitem{arXiv2022_temporal}
S.~Wang, X.~Zhang, G.~Zhang, Y.~Xiong, G.~Tian, S.~Guo, and J.~Li, ``Temporal
  and spatial online integrated calibration for camera and lidar,'' {\em arXiv
  preprint arXiv:2207.10454}, 2022.

\bibitem{sun2022atop}
Y.~Sun, J.~Li, Y.~Wang, X.~Xu, X.~Yang, and Z.~Sun, ``Atop: An
  attention-to-optimization approach for automatic lidar-camera calibration via
  cross-modal object matching,'' {\em IEEE Transactions on Intelligent
  Vehicles}, vol.~8, no.~1, pp.~696--708, 2022.

\bibitem{koide2023general}
K.~Koide, S.~Oishi, M.~Yokozuka, and A.~Banno, ``General, single-shot,
  target-less, and automatic lidar-camera extrinsic calibration toolbox,'' {\em
  arXiv preprint arXiv:2302.05094}, 2023.

\bibitem{schneider2017regnet}
N.~Schneider, F.~Piewak, C.~Stiller, and U.~Franke, ``Regnet: Multimodal sensor
  registration using deep neural networks,'' in {\em 2017 IEEE intelligent
  vehicles symposium (IV)}, pp.~1803--1810, IEEE, 2017.

\bibitem{iyer2018calibnet}
G.~Iyer, R.~K. Ram, J.~K. Murthy, and K.~M. Krishna, ``Calibnet: Geometrically
  supervised extrinsic calibration using 3d spatial transformer networks,'' in
  {\em 2018 IEEE/RSJ International Conference on Intelligent Robots and Systems
  (IROS)}, pp.~1110--1117, IEEE, 2018.

\bibitem{shan2018lego}
T.~Shan and B.~Englot, ``Lego-loam: Lightweight and ground-optimized lidar
  odometry and mapping on variable terrain,'' in {\em 2018 IEEE/RSJ
  International Conference on Intelligent Robots and Systems (IROS)},
  pp.~4758--4765, IEEE, 2018.

\bibitem{RANSAC}
M.~A. Fischler and R.~C. Bolles, ``Random sample consensus: a paradigm for
  model fitting with applications to image analysis and automated
  cartography,'' {\em Communications of the ACM}, vol.~24, no.~6, pp.~381--395,
  1981.

\bibitem{canny1986computational}
J.~Canny, ``A computational approach to edge detection,'' {\em IEEE
  Transactions on pattern analysis and machine intelligence}, no.~6,
  pp.~679--698, 1986.

\bibitem{huber1992robust}
P.~J. Huber, ``Robust estimation of a location parameter,'' in {\em
  Breakthroughs in statistics}, pp.~492--518, Springer, 1992.

\bibitem{vkitti2}
Y.~Cabon, N.~Murray, and M.~Humenberger, ``Virtual kitti 2,'' 2020.

\bibitem{he2016deep}
K.~He, X.~Zhang, S.~Ren, and J.~Sun, ``Deep residual learning for image
  recognition,'' in {\em Proceedings of the IEEE conference on computer vision
  and pattern recognition}, pp.~770--778, 2016.

\bibitem{deng2009imagenet}
J.~Deng, W.~Dong, R.~Socher, L.-J. Li, K.~Li, and L.~Fei-Fei, ``Imagenet: A
  large-scale hierarchical image database,'' in {\em 2009 IEEE conference on
  computer vision and pattern recognition}, pp.~248--255, Ieee, 2009.

\bibitem{kingma2014adam}
D.~P. Kingma and J.~Ba, ``Adam: A method for stochastic optimization,'' {\em
  arXiv preprint arXiv:1412.6980}, 2014.

\bibitem{Ceres2022}
S.~Agarwal, K.~Mierle, and T.~C.~S. Team, ``{Ceres Solver},'' 3 2022.

\bibitem{kitti}
A.~Geiger, P.~Lenz, C.~Stiller, and R.~Urtasun, ``Vision meets robotics: The
  kitti dataset,'' {\em International Journal of Robotics Research (IJRR)},
  2013.

\bibitem{yuan2020rggnet}
K.~Yuan, Z.~Guo, and Z.~J. Wang, ``Rggnet: Tolerance aware lidar-camera online
  calibration with geometric deep learning and generative model,'' {\em IEEE
  Robotics and Automation Letters}, vol.~5, no.~4, pp.~6956--6963, 2020.

\end{thebibliography}

\end{document}